# Adapted sampling for 3D X-ray computed tomography

Anthony Cazasnoves and Fanny Buyens
CEA, LIST,
91191 Gif-sur-Yvette
Email: anthony.cazasnoves@cea.fr

Sylvie Sevestre
CEA, LIST, 91191 Gif-sur-Yvette
MAP5, Universite Paris Descartes, Paris Cedex06, France 75270
Email: sylvie.ghalila@cea.fr

*Abstract*—In this paper, we introduce a method to build an adapted mesh representation of a 3D object for X-Ray tomography reconstruction. Using this representation, we provide means to reduce the computational cost of reconstruction by way of iterative algorithms. The adapted sampling of the reconstruction space is directly obtained from the projection dataset and prior to any reconstruction. It is built following two stages : firstly, 2D structural information is extracted from the projection images and is secondly merged in 3D to obtain a 3D pointcloud sampling the interfaces of the object. A relevant mesh is then built from this cloud by way of tetrahedralization. Critical parameters selections have been automated through a statistical framework, thus avoiding dependence on users expertise. Applying this approach on geometrical shapes and on a 3D Shepp-Logan phantom, we show the relevance of such a sampling - obtained in a few seconds - and the drastic decrease in cells number to be estimated during reconstruction when compared to the usual regular voxel lattice. A first iterative reconstruction of the Shepp-Logan using this kind of sampling shows the relevant advantages in terms of low dose or sparse acquisition sampling contexts. The method can also prove useful for other applications such as finite element method computations.

## I. INTRODUCTION

Computed tomography (CT) scanners data are basically reconstructed with the gold standard Feldkamp (FDK) algorithm [1]. Its analytical formalism requires a large number of projections for a robust reconstruction thus not contributing to limit dose to patient. Calling on iterative methods - such as ART [2], EM [3] and their derivatives - reliable reconstructions are performed from reduced dataset but at large computational cost. This is mainly due to the regular voxel lattice used as sampling : high spatial resolution means thin 3D grids and leads to an oversampling of large homogeneous regions. This translates to a larger number of unknowns to estimate - computational cost - and to large file to store the reconstruction - memory requirement. Graphics processing units (GPU) downscale iterative methods computation time but the processing of big volumes still remains an issue - due to limitation in devices internal memory size. Volume storage memory consumption moreover remains the same.

Addressing this issue, representations enabling an adaptive sampling of the reconstruction volume have been investigated. Such is the case of multi-scale basis functions and especially of blobs [4]. Main drawbacks are however their high computational cost and the complexity of extension to the 3D case. In tomography, meshed representations are of particular interest, due to their ability to achieve a sampling mirroring the structure of the object. Meshed 2D CT reconstruction was investigated in [5] and the 3D case in [6], [7]. Brankov *et al.* [5] approach is of sampling nature : a pixel-based coarse reconstruction is first performed and is used to sample the mesh nodes adequately. A maximum-likelihood (ML) algorithm adapted to the mesh representation is then applied for reconstruction. The initial pixel reconstruction needed to build the triangulated representation of the object is the main limitation of this approach because of the large number of projections required by analytical algorithms and because it represents an additional step. In [6], Sitek *et al.* introduce a method based on a refinement scheme. A first regular grid of tetrahedral cells is generated and several iterations of EM algorithm are performed. Tetrahedra linked to big errors are splitted by addition of a node at their centroids and EM is again performed. In [7] the approach is of coarsening type. Starting with a fine grid of tetrahedra, the method proceeds alternating iterative reconstruction and collapsing of cells belonging to homogeneous regions. In both cases, remeshing operations are computationally costly and the number of nodes added at each iteration being user fixed, the performance will be linked to the one's expertise. Buyens *et al.* [8] framework combines the idea of the previous approach. Reconstruction is first performed on a 2D grid of triangle. By interpolating the result to a pixel grid, a level-set method is used to re-sample the nodes and a more adapted mesh is generated. Values are interpolated back from the pixel base to the mesh one and the process goes through another iteration. Results show that the convergence of the reconstruction is substantially improved when the mesh matches the structure of the considered object. The issue is once again that the representation adapts itself to the object along with the tomographic reconstruction. Moreover, the values interpolation from the tessellated representation to the pixel grid and back prove to be costly in terms of computation and may introduce imprecision in the reconstruction.

In this work we build an adapted mesh prior to any step of reconstruction by directly exploiting the acquired data. Doing so, fast 3D CBCT reconstructions are achievable. In order to create such a mesh the location of the 3D interfaces that constitute the object structure has to be known. Firstly, we exploit the evidence of 2D interfaces as the result of the 3D ones by performing edge detection on the acquired data. Secondly, the 2D structural information is merged in 3D using the statistical framework of the hypothesis testing.

This papers is organized as follows. Section II is devoted to the structural information merging and the positioning of the mesh nodes as a pointcloud. Section III shows the results of the complete method applied on numerical data. Conclusions and perspectives of this work are discussed in Section IV.

## II. AUTOMATED 3D POINTCLOUD SAMPLING OF THE RECONSTRUCTION VOLUME

The mesh representation could be well adapted to the object in reconstruction when the mesh nodes are located on the 3D object interfaces. The aim hence is to obtain a 3D pointcloud adequately sampling these interfaces. We estimate these 3D locations in two steps by exploiting the evidence of 2D interfaces as a result of 3D ones. As a first step, we extract the structural information of the 2D projections by performing edge detection using Canny's filter [9]. The second step consists in the fusion of the detected 2D edge maps - Fig 1(a) - performed by successively:

- Merging by accumulation within the 3D volume
- Filtering the resulting scalar field to only keep the relevant information and place the nodes accordingly

The merging is carried through standard ray-driven backprojection on a 3D regular grid of voxels. Calling on the statistical framework, the filtering step is fully automated and accurate, thus not depending on the user's expertise. By placing nodes with respect to this filtering, a 3D pointcloud sampling of the volume is built - Fig 1(c). Sparse outliers removal based on a K-nearest neighbour approach [10] is then performed. The final mesh is obtained from the cloud using Tetgen algorithm [11]- Fig 1(d).

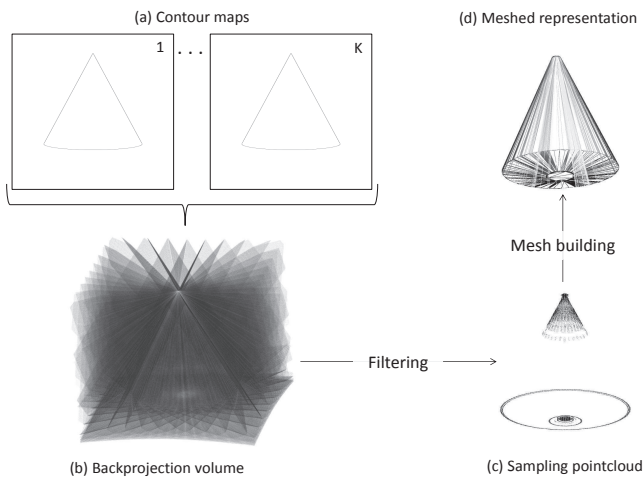

Fig. 1: (a) Examples of 2D edge maps. (b) Backprojection over the 3D volume (white corresponds to a null count). (c) 3D pointcloud sampling of the volume obtained by filtering the count volume. (d) Mesh built on pointcloud (c).

To achieve the filtering, at each $l^{th}$ voxel of the volume, we examine the count value $n_l$ resulting of the backprojection of the $K$ 2D edge images as

$$n_l = \sum_{k=1}^{K} B_k^l \, , \, l \in \{1, \dots, M\}, \quad (1)$$

where $M$ is the total number of voxels in the grid and $B_k^l$ denotes the binary value of pixel in the $k^{th}$ edge image linked to voxel $l$ by backprojection.

Therefore, the value $n_l$ of voxel $l$ is the number of pixels identified as edge in the acquired images linked to $l$ by projection, thus constituting a first counting volume $(n_l)_{l=1,\dots,M}$ - Fig 1(b) - of interfaces.

To select the small set of relevant points for the mesh, one can put aside most common low count voxels of the volume. This amounts to defining the upper tolerant limit of the volume counts. As the resulting 3D scalar field $(n_l)$ is obtained by a counting phenomenon, it can therefore be modeled as a Poisson distribution. However, the 2D edge images are parsimonious and a majority of the voxels are assigned a null value by backprojection. To select relevant voxels among non null counts, we hence need to use more specific distribution model : the zero-truncated Poisson distribution (ZTP).

ZTP is characterized by a single parameter $\theta$ usually estimated by using Plackett's method [12]. Lets denote $\hat{\theta}$ the Plackett estimation using non null counts of the volume. As the count dataset is very large, one can consider $\theta = \hat{\theta}$. Therefore, the $L$ non null count voxels ($L < M$) of the volume, $(n_l)_{l=1,\dots,L}$ is a sample of a random variable $N$ where

$$N \sim ZTP(\theta). \quad (2)$$

If $\alpha$ denotes the confidence level of the upper tolerant limit $\lambda$, therefore

$$P(N \leq \lambda) = 1 - \alpha_{ZTP}. \quad (3)$$

The threshold $\lambda$ hence is by definition the quantile $Q$ of the ZTP distribution of significance level $1 - \alpha_{ZTP}$. We get $Q$ using the approximation of Gilchrist [13]. Letting $F(.)$ be the cumulative distribution function of the Poisson distribution of parameter $\theta$ one can find the equivalent confidence level $\alpha_P$ for the standard Poisson distribution to the required $\alpha_{ZTP}$ of the ZTP using

$$1 - \alpha_P = F(1) - (1 - \alpha_{ZTP})(1 - F(1)). \quad (4)$$

The required quantile is then straightforwardly given as $\lambda = Q_P(1-\alpha_P)$ with $Q_P$ the $\alpha_p$ quantile of the standard Poisson distribution. The set of centroids of voxels for which counts are larger than $\lambda$ thus defines the sampling 3D pointcloud on which the adapted mesh is built.

For specific configurations, this approach can produce artifacts in the pointcloud as planes filled with misplaced points. This happens when objects oriented orthogonally to the rotation axis and with particular aspect ratio are present in the volume.

It is due to the fact that acquisitions are limited to the $2\pi$-plane of rotation around the object of interest and that few projections are used to perform the pointcloud construction. In such configurations, the count distribution of $N$ is closer to Poisson law than to ZTP one. To take into account this variation, Mizere et al. [14] test is applied at each slice in order to decide between ZTP and Poisson law as a model for $N$. The decision is based on the following statistic

$$T_f = S \times \frac{V_S^2}{\overline{N_S}} \quad (5)$$

where $S$ is the number of voxels in the slice, $\overline{N_S}$ is the mean of its distribution and $V_S^2$ its unbiased empirical variance. The authors showed that the statistic $T_f \sim \chi^2_{S-1}$ when $N \sim \text{Pois}(\theta)$. We therefore choose the Poisson distribution when the statistic $T_f$ is superior to the $\alpha$ quantile of the Chi-square distribution with $S-1$ degree of freedom and $\alpha$ is the confidence level chosen for this statistical test.

### III. RESULTS

The method performance is evaluated on a 3 mono-material shapes - sphere, cone and bone extremity - and on the multi-material 3D phantom of Shepp-Logan. We simulate 30 projections - sizing $1024^2$ pixels - of these objects as dataset from which the meshes are built.

Pointcloud quality is based on its comparison to the STL phantom of the object. Using CloudCompare [15] we compute the closest distance $d_i$ from each point $i$ to the STL. Point selection is considered as optimum when $d_i \leq Grid_{res} \times \sqrt{3}/2$. Meshes are compared to the best voxel-based representations of the objects for $128^3$, $256^3$ and $512^3$ grids resolution. These descriptions are obtained using Binvox [16] for the three shapes. For the Shepp-Logan, this corresponds to a $512^3$ phantom. Confidence levels have to be chosen but the set of values is extremely reduced and well defined $\alpha \in \{0.5, 0.1, 0.001\}$. We choose $\alpha = 0.05$ for $128^3$ grids for the 3 shapes. At $256^3$ grid resolution, $\alpha = 0.01$ for both the sphere and the bone and it is set to $0.001$ for the cone. For the $512^3$ grid, $\alpha = 0.001$ for the cone and the bone and it is set to $0.01$ for the sphere. The clouds displayed in Fig 2(a)-(b) shows how adequately the pointclouds sample the object of interest. The quality metric - observation is confirmed by the quality metric displayed in Fig 2(c). Details of cuts in the sphere meshes superimposed with its best voxel description - Fig 2(d) - show how the mesh enables an extremely fine sampling along the interfaces of the object while at the same time displaying large cells in homogeneous areas. Considering the number of cells composing the mesh - Table I - its is clear that this fine description still yields an important reduction in the number of unknowns to estimate - up to 99%. Apart from the large cells used in homogeneous portions of the volume, this gain also comes from the fact that the reconstruction is restricted to the object convex hull.

Fig 3(a) shows that results are similar for the Shepp-Logan phantom. Nodes density - hence smaller cells - is concentrated along the borders of the ovoids defining the geometry. This is especially visible on the outer layer of the phantom. Eve though a the reduction in the number of unknowns is very important - 5 408 212 cells in the mesh, 4% of the equivalent grid - the close-up details show that the smallest volumes of the phantom are described with precision by our mesh. Again, a reduction in the number of unknowns to evaluate is achieved and the computational burden is focused on the areas of interest. A first version of an adapted SART algorithm was applied on this mesh - Fig 3(b) - providing an initial rough reconstruction of the volume.

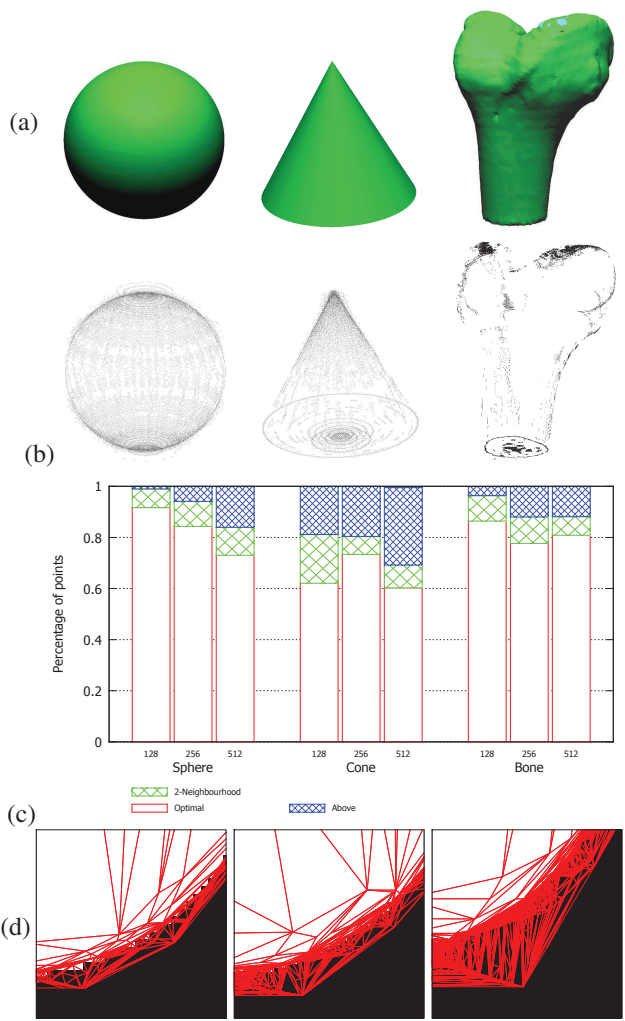

Fig. 2: (a)STL description of the three studied shapes. (b)Examples of corresponding pointclouds obtained by the method. (c)Clouds qualities for 3 grids resolution. (d) Zoom-in on a cut in the sphere meshed volume superimposed with best voxel description for $128^3$, $256^3$ and $512^3$ grids

### IV. CONCLUSION

This paper presents a reliable and automated method to build a 3D adapted mesh sampling of an object from a few number of 2D projections. A keypoint of this approach is the ability to obtain this representation prior to any usual tomographic reconstruction. The sampling is obtained by exploiting the evidence of 2D interfaces as result of 3D ones. Thus, by extracting edges present in the acquired raw data, a 3D merging scheme relying on a statistical model of the backprojection volume was developed to obtain a node clouds fitting the object interfaces. Using standard constraint Delaunay tetrahedralization, this cloud provides a content-adapted mesh for low computational burden reconstruction. The statistical flavour of the method enables critical parameters automatic choices thus avoiding results quality dependence on users expertise.

| Grid resolution | $128^3$ | $256^3$ | $512^3$ |
|---|---|---|---|
| Mesh cells | 18950 | 167925 | 1403775 |
| Ratio mesh/voxel | 0.9% | 1% | 1% |
| Total computation (s) | 4 | 15.6 | 32 |

TABLE I: Compression obtained by the mesh description of the sphere for 3 regular grids resolutions. Setup : Intel Xeon E31245 (3.3GHz), 8Gb of RAM and a Tesla C2070

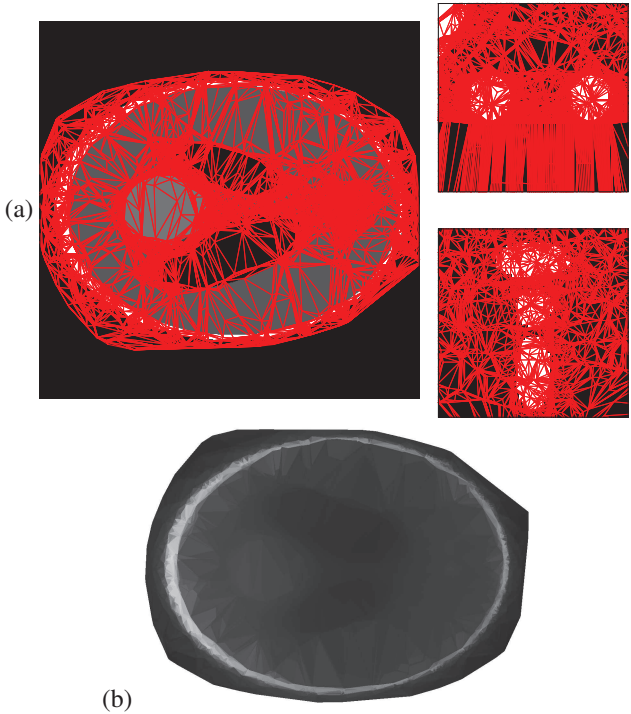

Fig. 3: (a) Left : Superposition of the transverse cut in the meshed volume and of the corresponding voxel slice of the Shepp-Logan phantom. Right : Zoom-in of the mesh in the sagittal and transverse cut. (b) Transverse cut in the volume reconstructed using a mesh adapted SART.

Using extremely reduced dataset - 30 projections - our method provides reliable adapted mesh sampling of the 3D object in a matters of seconds on a conventional setup. In terms of automation, the choice of $\alpha$ levels are still up to the user but from an extremely reduced set. As illustrated in Section III, the reduction in the number of cells used in the description is very important - up to 99% - and the method manages to focus the cells density around the interfaces of the various volumes composing the object - thus ensuring its good geometrical description even with so few cells.

Considering the dose exposure to the patient, this new type of sampling can prove interesting. Reliable representations are obtained from very sparse dataset and a first rough reconstruction was obtained through an adapted SART using a suitable projector/backprojector [17]. Our work will now focus on the improvement of the reconstruction by investigating a better adaption of the iterative algorithm to this new representation. Focusing on the pointcloud obtained by our method, one can think on obtaining a surface representation of the object of interest from it. Such surface modeling could be use in multiple medical applications such as, for example, the design of patient-specific prosthetics.